%% file: main.tex
\documentclass{article}

\usepackage{PRIMEarxiv}

\usepackage[utf8]{inputenc}
\usepackage[T1]{fontenc}
\usepackage{hyperref}
\usepackage{url}
\usepackage{booktabs}
\usepackage{amsfonts}
\usepackage{nicefrac}
\usepackage{microtype}
\usepackage{lipsum}
\usepackage{fancyhdr}
\usepackage{graphicx}
\graphicspath{{media/}}
\usepackage{siunitx}
\usepackage{natbib}
\usepackage{pgfplots}
\usepackage{tikz}
\usetikzlibrary{patterns}
\usepackage{caption}
\usepackage{float}
\usepackage{amsmath}
\usepackage{bm}
\usepackage{adjustbox}
\usepackage{pgffor}
\usepackage{enumitem}

\let\ENDGRAPH\endinput

\usepackage[capitalise]{cleveref}
  
\title{Falcon2-11B Technical Report}

\author{
\textbf{Quentin Malartic}$^*$ \and \textbf{Nilabhra Roy Chowdhury} \and \textbf{Ruxandra Cojocaru} \and \textbf{Mugariya Farooq} \\
\and \textbf{Giulia Campesan} \and \textbf{Yasser Abdelaziz Dahou Djilali} \and \textbf{Sanath Narayan} \and \textbf{Ankit Singh}\\
\and \textbf{Maksim Velikanov} \and \textbf{Basma El Amel Boussaha} \and \textbf{Mohammed Al-Yafeai}\and \textbf{Hamza Alobeidli} \\
\and \textbf{Leen Al Qadi} \and \textbf{Mohamed El Amine Seddik} \and \textbf{Kirill Fedyanin} \and \textbf{Reda Alami} \\  \and \textbf{Hakim Hacid} \\\\
  Technology Innovation Institute, Abu Dhabi, United Arab Emirates \\\\
  \texttt{$^*$Falcon-LLM[at]tii[dot]ae} \\
}

\definecolor{adaptation}{HTML}{FCCDE5}
\definecolor{objective}{HTML}{FFC2BA}
\definecolor{architecture}{HTML}{FaF9B3}
\definecolor{evaluation}{HTML}{B9DEFF}
\definecolor{neutral}{HTML}{CFCFCF}
\definecolor{fine-tuning}{HTML}{D7F3E7}
\definecolor{fine-tuning-darker}{HTML}{c4eddc}
\pgfplotsset{compat=1.18}
\begin{document}
\maketitle

\begin{abstract}
We introduce Falcon2-11B, a foundation model trained on over five trillion tokens, and its multimodal counterpart, Falcon2-11B-vlm, which is a vision-to-text model. We report our findings during the training of the Falcon2-11B which follows a multi-stage approach where the early stages are distinguished by their context length and a final stage where we use a curated, high-quality dataset. Additionally, we report the effect of doubling the batch size mid-training and how training loss spikes are affected by the learning rate. The downstream performance of the foundation model is evaluated on established benchmarks, including multilingual and code datasets. The foundation model shows strong generalization across all the tasks which makes it suitable for downstream finetuning use cases. For the vision language model, we report the performance on several benchmarks and show that our model achieves a higher average score compared to open-source models of similar size. The model weights and code of both Falcon2-11B and Falcon2-11B-vlm are made available under a permissive license.
\end{abstract}

\section{Introduction}

The first generation of Falcon models, featuring Falcon-7B, Falcon-40B, and Falcon-180B \citep{falcon}, made a significant contribution to the open-source community, promoting the release of advanced LLMs with permissive licenses. In this report, we introduce a second generation of models, Falcon2, focused on increased usability and integrability, towards building a multi-modal ecosystem currently composed of a large language model with 11B parameters and a corresponding vision language model.

Historically, large language models first saw an important rise in performance with increased model size \citep{gpt3, palm}. Updated scaling laws \citep{chinchilla} brought to light that this initial generation of large language models were most likely undertrained, highlighting the need for more training data to further increase the performance. This triggered another important paradigm shift, namely moving from large curated datasets \citep{pile, palm}, to large-scale datasets harvesting mostly web data from the CommonCrawl project, such as RefinedWeb \citep{refinedweb} or RedPajama \citep{redpajama}. Both these advances led to the release of large open-source models such as Llama-65B \citep{llama} and Falcon-180B \citep{falcon}. More recently, the Llama2 models \citep{llama2} showed the benefits of even more prolonged training, achieving state-of-the-art performance with smaller model sizes. This trend was followed in the past year, resulting in a number of small-sized yet highly performing models such as Qwen-7B \citep{qwen}, Mistral-7B \citep{mistral}, Yi-6B and Yi-9B \citep{yi}, Gemma-7B \citep{gemma} and Llama3-8B \citep{llama3}.

We present the Falcon2-11B LLM model \footnote{\url{https://huggingface.co/tiiuae/falcon-11B}}, achieving better performance on the Open LLM Leaderboard \footnote{\url{https://huggingface.co/spaces/open-llm-leaderboard/open_llm_leaderboard}} tasks compared to Mistral-7B and Llama3-8B, and similar to Gemma-7B. The Falcon2-11B model surpasses the performance of the larger Falcon-40B, ensuring reduced computational cost and faster inference for end users while conserving its multilingual aspect. We also present the Falcon-11B-vlm \footnote{\url{https://huggingface.co/tiiuae/falcon-11B-vlm}}, a vision language model built from the Falcon2-11B foundation model. Both models are available under a permissive open-source license as detailed in \cref{sec:license}.

\cref{sec:pretraining} offers pre-training details for the Falcon2-11B LLM, including model architecture, training hyperparameters and the effect of batch size doubling, training stages to enable context length extension and multilingual support, while \cref{subsec:eval} presents the model's performance across multiple benchmarks, including multilingual and code datasets. \cref{sec:vlm} presents the multimodal extension to a vision languages model, Falcon2-11B-vlm, including training details and evaluation results. Model availability and license information are covered in \cref{sec:license}.

\section{Pre-training}\label{sec:pretraining}

\subsection{Architecture}\label{sec:architecture}
Falcon2-11B is based on the transformer architecture \citet{vaswani2017attention}. As in the original Falcon models \citet{falcon}, we use grouped query attention (GQA) \citep{multiquery} with $n_\text{kv}=8$. Additionally, the model was trained using a tensor parallel distribution $\text{TP}=8$. Similarly to \citet{palm-2} and \citet {falcon}, we use parallel transformer blocks, with the multilayer perceptron (MLP) and self-attention blocks positioned in parallel, and with a single layer norm (LN) per block. More specifically, the typical transformer block can be written as
\begin{subequations}
\begin{align}
    x_{\mathrm{attn}} &= \text{LayerNorm}_\mathrm{attn}(x + \text{Attention}(x)),\\
    \text{Block}(x) &= \text{LayerNorm}_\mathrm{mlp}(\mathrm{MLP}(x_{\mathrm{attn}}) + x_{\mathrm{attn}}),
\end{align}
\end{subequations}
while all presented Falcon\footnote{Note that Falcon2-11B and Falcon-7B use the presented implementation, while Falcon-40B and Falcon-180B use a different layernorm for the attention and the MLP inputs.} models use the parallel transformer block, expressed as:
\begin{subequations}
\begin{align}
    x_\mathrm{norm} &= \text{LayerNorm}(x),\\
    \text{ParallelBlock}(x) &= x + \text{MLP}(x_\mathrm{norm}) + \text{Attention}(x_\mathrm{norm}).
\end{align} 
\end{subequations}
The summary of the Falcon2-11B architecture, as well as that of the previous Falcon models is presented in \cref{tab:architecture}. The model is rather deep, with the same number of layers as the Falcon-40B. It has been designed so that one can deploy it on a single A10 GPU (24 GB memory). The main differences are
\begin{itemize}[topsep=0pt,itemsep=0pt,parsep=5pt,partopsep=0pt]
    \item A larger head dimension, that leads to a slight boost in performances at unreported 3B scale experiments as well as a reduced memory footprint during training, especially at longer context window;
    \item An increased $\theta_\mathrm{rope}$ for long context, following the results reported by \citet{meta_long_context,llama_code}.
\end{itemize}

\begin{table}[ht]
 \caption{Summary of the Falcon models architectures.}
  \centering
  \scriptsize
  \adjustbox{width=\columnwidth}{
  \begin{tabular*}{0.98\linewidth}{lccccccccc}
    \toprule
    \textbf{Model} &  \texttt{n\_layers} & \texttt{d\_model} &\texttt{n\_heads} & \texttt{head\_dim} & \texttt{n\_kv} & \texttt{context\_length} & \texttt{rope\_base} & \texttt{attention} & \texttt{tied\_emb} \\
    \midrule
    Falcon-7B   & 32 & 4544  & 71  & 64  & 1  & 2048      & 10K & FA & Yes \\
    Falcon-40B  & 60 & 8192  & 128 & 64  & 8  & 2048      & 10K & FA & Yes \\
    Falcon-180B & 80 & 14848 & 232 & 64  & 8  & 2048      & 10K & FA & Yes \\
    \midrule
    Falcon2-11B \textit{stage 1} & 60 & 4096   & 32  & 128 & 8 & 2048 & 5M+42 & FA2 & Yes \\
    Falcon2-11B \textit{stage 2} & {\color{gray}60} & {\color{gray}4096} & {\color{gray}32} & {\color{gray}128} & {\color{gray}8} & 4096 & {\color{gray}5M+42} & {\color{gray}FA2} & {\color{gray}Yes}\\
    Falcon2-11B \textit{stage 3} & {\color{gray}60} & {\color{gray}4096} & {\color{gray}32} & {\color{gray}128} & {\color{gray}8} & 8192 & {\color{gray}5M+42} & {\color{gray}FA2} & {\color{gray}Yes}\\
    Falcon2-11B \textit{stage 4} & {\color{gray}60} & {\color{gray}4096} & {\color{gray}32} & {\color{gray}128} & {\color{gray}8} & {\color{gray}8192} & 500K+42 & {\color{gray}FA2} & No \\
    \bottomrule
  \end{tabular*}}
  \label{tab:architecture}
\end{table}

As explained in \cref{sec:data,sec:training}, the training was split into four stages. During the first three stages, tied embeddings were used, similar to the first generation of Falcon models \citep{falcon}. For \textit{stage 4}, we chose to untie the embeddings from the projector weights as it lead to better performances, increasing the number of trainable model parameters ($\approx 300M$).

\paragraph{FlashAttention} 
To maximize GPU utilization, we added support for FlashAttention-2 (FA2) \citep{flash_attn_2} to our training routine. FA2 is reported to be at least 1.3 times faster than its predecessor FlashAttention (FA) and 10 times faster than the native PyTorch implementation. We note that the speed-up only applies to the attention layer in each Transformer block. As a result, the total speed gain for training a model is far less than 1.3x for shorter sequences of 2K. For longer sequence lengths, of 4K and 8K, the benefits of FA2 are more profound as the compute needed for the attention calculation far exceeds the compute needed for the MLP layers, reaching a speedup of up to 50\% for the 8K context window.

The open-source Python implementation of FA2 also allowed us to use 128-dimensional attention heads. As expected, we observed this to increase throughput at high context window when compared with 64-dimensional attention heads. Additionally, we also observe a slight boost in downstream performances on smaller scales. 

\subsection{Dataset}\label{sec:data} 
Our data pipeline is largely based on \citet{refinedweb, falcon}. We consider two kinds of data sources:
\begin{itemize}[topsep=0pt,itemsep=0pt,parsep=5pt,partopsep=0pt]
    \item Extensively processed web data (RefinedWeb), in the following natural languages: English, German, Spanish, French, Italian, Dutch, Polish, Portuguese, Czech, Romanian, and Swedish;
    \item Curated corpora, including code, conversations (general and technical), books, scientific publications and technical reports, encyclopedias, and patents.
\end{itemize}
\paragraph{Quality filtering} One of the limitations of the RefinedWeb dataset from \citet{refinedweb} was that the manually tuned quality heuristics were developed for English exclusively, resulting in sub-optimal multilingual data quality. For the Falcon2-11B training, we adapted the RefinedWeb filtering rules to multilingual data. More specifically, we manually adapted the matching patterns for line quality filtering, as well as the stop words. We provide more details on the filtering rules we used for multilingual data in \cref{sec:filter_details}. The filtering rates for each non-English language are reported in \cref{tab:multilingual_filtering}.

\begin{table}[ht]
 \caption{Multilingual data sources quality filtering rates (\%). Removal rates are with respect to the last pre-processing step used for multilingual data in \cite{falcon}, to reflect the effect of the new filters. The extra quality heuristics filtering consists of language-tuned versions of average word length and minimum stop word filter thresholds originally used for English following \citet{gopher}.}
  \centering
\begin{tabular}{lccc}
\toprule
\textbf{Language} & \textbf{Line-wise filter (\%)} & \textbf{Extra heuristics (\%)} & \textbf{Total removed (\%)} \\
\midrule
German (\textit{de}) & 8.57 & 1.56 & 10.13 \\
Spanish (\textit{es}) & 6.6 & 3.37 & 9.97 \\
French (\textit{fr}) & 8.51 & 0.72 & 9.23 \\
Italian (\textit{it}) & 12.16 & 1.29 & 13.45 \\
Dutch (\textit{nl}) & 8.34 & 0.75 & 9.09 \\
Polish (\textit{pl}) & 6.2 & 1.17 & 7.37 \\
Portuguese (\textit{pt}) & 8.73 & 2.1 & 10.83 \\
Czech (\textit{cz}) & 9.5 & 0.85 & 10.35 \\
Romanian (\textit{ro}) & 5.25 & 1.23 & 6.48 \\
Swedish (\textit{sv}) & 5.56 & 1.4 & 6.96 \\
\bottomrule
\end{tabular}

  \label{tab:multilingual_filtering}
\end{table}

\paragraph{Conversation trees}
In \cite{falcon}, the conversation tree data (originating from Reddit, Stack Exchange, and Hackernews) was prepared by manipulating the attention mask as well as the positions. In this case, it was possible to train on a whole conversation tree without repeating data, only allowing a given conversation message to attend to its own trajectory. With the official implementation of the FA2 algorithm from \citet{flash_attn_2}, it was not possible to manipulate the attention mask. For this reason, we flatten each conversation tree into all of the possible threads, thus repeating each conversation message that has multiple replies. Upon repeating data, we mask the loss on the repeated occurrences, making the training process almost equivalent, up to the truncation of the longest conversation to fit within the context window when applicable. A visual representation of this process is given in \cref{fig:convosGraph}. We estimate the overall compute overhead due to data repetition to be $+\num{79}\%$ for Reddit, $+\num{164}\%$ for Stack Exchange, and $+\num{97}\%$ for Hacker News. In the least favorable configuration (\textit{stage 1}), these data sources represent respectively $\num{6,1}\%$, $\num{0,72}\%$ and $\num{0,04}\%$ of the training dataset, corresponding to an overall overhead of $+\num{6}\%$, which is compensated by the speed-up from FA2 of between 5.5\% and 50\%, depending on the size of the context window.

  \long\def\GRAPH #conversation_graphconvosGraph {}%
  \input{conversation_graph}

\paragraph{Code data}
In order to expand the downstream applications of Falcon2-11B, web-scraped code data has been included in the pre-training phase. Code samples have been extracted from the deduplicated version of The Stack dataset \citep{thestack}, featuring source code with permissive licenses in 358 programming languages, from which we select scripts in 43 programming languages (see \cref{sec:code_languages} for a full list). The selected samples have been pre-processed via natural language identification, using a \textit{fasttext} classifier \citep{fasttextlang}, retaining only those samples belonging to the eleven natural languages of interest,  with a language score above $0.15$. Afterward, a minimum word count of $50$, a minimum alphanumeric character threshold of $0.1$, and a maximum average word length of $100$ were applied, the latter to remove samples containing mainly key information and/or long URLs. Finally, exact substring deduplication was applied on top of the \textit{stack-deduped} dataset\footnote{\url{https://huggingface.co/datasets/bigcode/the-stack-dedup}}, following the methodology describer in \citet{refinedweb}.

\paragraph{Data sources mixture}
We train on a mixture of processed, filtered, and deduplicated English and multilingual web data, as well as a selection of curated corpora. We consider the following natural languages, by the rate of occurrence: English (\textit{en}), German (\textit{de}), Spanish (\textit{es}), French (\textit{fr}), Italian (\textit{it}), Dutch (\textit{nl}), Polish (\textit{pl}), Portuguese (\textit{pt}), Czech (\textit{cz}), Romanian (\textit{ro}) and Swedish (\textit{sw}). The curated corpora are composed of the following datasets:
\begin{itemize}[topsep=0pt,itemsep=0pt,parsep=5pt,partopsep=0pt]
    \item Code from The Stack \citep{thestack};
    \item Scientific articles in tex format from arXiv, and in plain text from PubMed;
    \item Conversations from Reddit, Stack Exchange, Hacker News;
    \item Books;
    \item Patents from USPTO.
\end{itemize}
To enable multilingual capabilities, we use a significant fraction (of $15\%$ or more) of non-English web data throughout the training, which is at least twice the amount used in the previous Falcon series. Consequently, we reduce the fraction of English web data while keeping the source-specific data rate within the same ballpark. Finally, in order to optimize the training and processing of long context samples, the training is divided into four stages:
\begin{itemize}[topsep=0pt,itemsep=0pt,parsep=5pt,partopsep=0pt]
    \item \textit{Stage 1} is performed on 4500 GT \footnote{We use the unit 1 GT to refer to $10^9$ tokens} with a context window of 2048;
    \item \textit{Stage 2} is performed on 250 GT with a context window of 4096;    
    \item \textit{Stage 3} is performed on 250 GT with a context window of 8192;
    \item \textit{Stage 4} is performed on 500 GT, training over multiple epochs of a high-quality proprietary mixture of general, technical, code, math, and reasoning data, with a context window of 8192.
\end{itemize}
 We detail in \cref{tab:data_mixtures} the data mixture of the Falcon2-11B first three stages, as well as that of the previous Falcon models for comparison.
\begin{table}[ht]
 \caption{Summary of the Falcon models training data mixture, including Falcon2-11B \textit{stages 1-3}. Unless specified otherwise, the multilingual data comprises a mixture of German, Spanish, French, Italian, Dutch, Polish, Portuguese, Czech, Romanian, and Swedish, ordered by their trained token rates from most to least. The source-specific corpora consist of code, books, conversation data, and technical data, ordered by their trained token rates from most to least.}
  \centering
  \begin{tabular}{lcccccc}
    \toprule
    \textbf{Model} & \multicolumn{2}{c}{\textbf{RefinedWeb}} & \multicolumn{2}{c}{\textbf{Source-specific}} & \textbf{Max length}  & \textbf{Tokens (GT)}\\
    \cmidrule(r){2-3} \cmidrule(r){4-5}
     & \textbf{English (\%)}    & \textbf{Multilingual (\%)} & \textbf{Code (\%)} & \textbf{Others (\%)} & \\
    \midrule
    Falcon-7B   & \num{79}   & \num{3} \textit{only fr}  & \num{3} & \num{15} & 2048  &  1500 \\
    Falcon-40B  & \num{74,6} & \num{7,5} & \num{5} & \num{12,9} & 2048  & 1000    \\
    Falcon-180B & \num{76,3} & \num{7,5} & \num{3,3} & \num{12,9} & 2048  & 3500\\
    \midrule
    Falcon2-11B \textit{stage 1} & \num{69,1} & \num{16,6} & \num{2,2} & \num{12,1} & 2048 & 4500 \\
    Falcon2-11B \textit{stage 2} & \num{57}   & \num{15,8} & \num{10,6} & \num{16,6} & 4096 & 250 \\
    Falcon2-11B \textit{stage 3} & \num{61}   & \num{15} & \num{10,5} & \num{13,5} & 8192 & 250 \\
    \bottomrule
  \end{tabular}
  \label{tab:data_mixtures}
\end{table}

\paragraph{Long context data}
With the aim of optimizing the training of Falcon2-11B, we made the decision to split it into four stages, a first stage with up to \num{2048} tokens per sample, a second stage with up to \num{4096} tokens per sample, and finally stage 3 and 4, with up to \num{8192} tokens per sample. To give some more details about the sample length distribution, in \textit{stage 2}, more than $\num{55}\%$ of trained tokens are part of samples with more than \num{2048} tokens. During \textit{stage 3}, more than $\num{71}\%$ of trained tokens are part of samples with more than \num{2048} tokens, and more than $\num{18}\%$ of trained tokens are part of samples with more than \num{4096} tokens. Because the sample length distribution can be quite different from one data source to another, we have adapted the data source mixtures from one stage to another, as reported in \cref{tab:data_mixtures}. Finally, to try and reduce the source distribution shifts between stages, some long samples across data sources have been broken down into smaller ones to be included in earlier stages.

\subsection{Training}\label{sec:training} 

\paragraph{Setup} The Falcon2-11B model was trained with model parameters and gradients in \texttt{bfloat16} precision, using the AdamW optimizer with $\beta_1=0.9$, $\beta_2=0.999$ and $\epsilon = 10^{-7}$, and storing the optimizer state (parameters, accumulated gradients and exponential moving averages) in \texttt{float32} precision. In terms of learning rate schedule, \textit{stage 1} started with a linear learning rate warm-up over the first 4 GT up to the maximum value of $\eta_\mathrm{max}=3.7\times 10^{-4}$,
followed by a cosine decay down to the minimum learning rate value of $\eta_\mathrm{min}=1.89\times 10^{-5}$. \textit{Stage 2-4} used a constant learning rate of $1.89\times 10^{-5}$. A weight decay value of $0.1$ and Z-loss of $10^{-4}$ were used, similarly to previous Falcon models. 

\paragraph{Context length increase} \textit{Stages 1-3} used a sequence length of $2048$, $4096$ and $8192$ respectively, with $\theta_{\mathrm{RoPE}} = 5M+42$. During \textit{stage 4}, the sequence length was maintained at $8192$, whereas $\theta_{\mathrm{RoPE}}$ was decreased to $500K+42$. The use of FA2 ensured a good throughput as the sequence length was increased, as shown in \cref{tab:throughput}.

\paragraph{Spike management} Due to the architecture constraints described in Section \ref{sec:architecture}, Falcon2-11B presents a deeper architecture given its hidden dimension size; therefore, spikes were occasionally experienced throughout its training. Indeed, the large depth limit can induce training instability through exploding gradients as proven in \citet{hayou2021stable}. In practice, spike management consisted of rollbacks and data skips as detailed in \citet{palm}. Moreover and in line with \citet{spike-theory}, we observed a reduction in spike frequency as the learning rate decayed.

\begin{figure}
    \centering
    \includegraphics{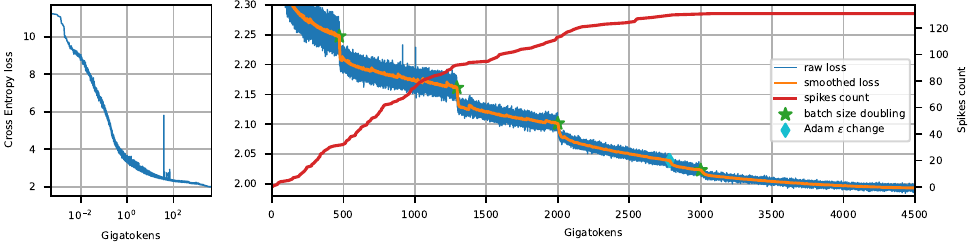}
    \caption{\textit{Stage 1} training: cross-entropy loss of next token prediction in logarithm (left) and linear (right) scale of trained tokens. The right plot also shows a cumulative count of spikes which caused a rollback and restart of training. Each of the 4 stars depicts a doubling of batch size, resulting in a history of $2K \rightarrow 4K \rightarrow 8K \rightarrow 16K \rightarrow 32K$ of the number of samples in the batch. Presumably, batch size doublings helped to reduce the (initially large) frequency of spikes. Finally, the diamond depicts the change of Adam $\varepsilon$ parameter from $10^{-8}$ to $10^{-7}$.}
    \label{fig:loss_curve}
\end{figure}

\paragraph{Batch size schedule} To reduce the number of spikes, we reduced the noise in the gradients by doubling the batch size at around 470 GT. As shown in \cref{fig:loss_curve}, the increase in batch size resulted in an abrupt drop in the training loss. Motivated by this drop, we doubled the batch size three more times during training \textit{stage 1}, resulting in a total increase from the initial batch size value $B=2048$ to the final $B=32768$. The origin of the sharp drop in loss after doubling the batch size $B$ is similar to the decay of learning rate $\eta$. In both cases, the loss decreases due to the removal of the noise from the optimization trajectory, with the noise magnitude being given by its temperature $T=\tfrac{\eta}{\sqrt{B}}$ \cite{malladi2022on}.

\begin{table}[ht]
 \caption{Falcon2-11B training throughput during \textit{stages 1-3}. nMT/H refers to the throughput per GPU.} 
  \centering
  \begin{tabular}{lcccccccccc}
    \toprule
    \textbf{Model}  & \texttt{context\_length}     & \textbf{DP}  & \textbf{PP} & \textbf{TP} & \textbf{Nodes} & \textbf{A100} & \textbf{Time} & \textbf{GT/H} & \textbf{nMT/H} & \textbf{TFLOPs} \\
    \midrule
    Falcon2-11B \textit{stage 1} & 2048 & 128 & 1 & 8 & 128 & 1024 & \SI{24}{\day} & \num{7,8} & \num{7,6} & \num{151}\\
    Falcon2-11B \textit{stage 2} & 4096 & 160 & 1 & 8 & 160 & 1280 & \SI{30}{\hour} & \num{8,5} & \num{6,6} & \num{146}\\
    Falcon2-11B \textit{stage 3} & 8192 & 160 & 1 & 8 & 160 & 1280 & \SI{36}{\hour} & \num{7,0} & \num{5,5} & \num{135}\\
    \bottomrule
  \end{tabular}
  \label{tab:throughput}
\end{table}

\begin{figure}
    \centering
    \includegraphics{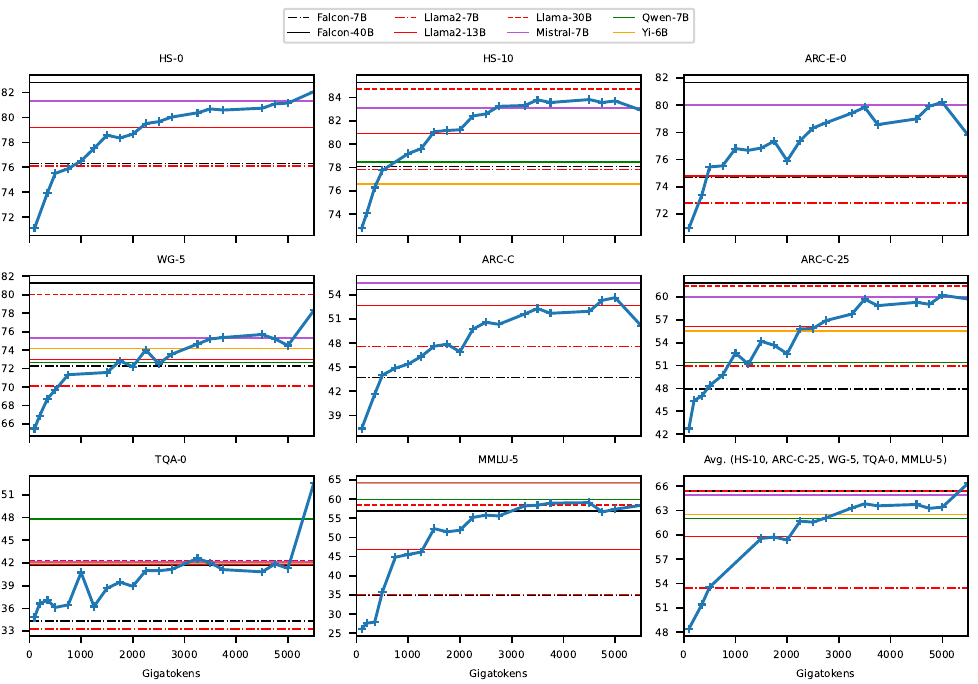}
    \caption{Evaluation results across \textit{stages 1-3}. We present the progression of evaluation results on English NLP tasks across different training checkpoints in \textit{stages 1-3}. Final results from the following models are given as reference: Falcon-7B, Falcon-40B, Llama2-7B, Llama2-13B, Llama-30B, Qwen-7B, and Yi-6B.}
    \label{fig:eval_progression}
\end{figure}

\subsection{Evaluation and Results}\label{subsec:eval}

\paragraph{English tasks} Throughout the training, we evaluated Falcon2-11B regularly, and report here the performances at 500 GT, 1000 GT, 1500 GT, 2000 GT, 2500 GT, 3000 GT, 3500 GT, 4500 GT, 4750 GT, and 5000 GT. For each of these checkpoints, we consider the maximum between the accuracy and accuracy of normalized completions of the following tasks: HellaSwag \citet{zellers2019hellaswag}, HellaSwag 10-shots \citet{zellers2019hellaswag}, AI2 Reasoning Challenge Easy \citet{allenai:arc}, Winogrande \citet{DBLP:journals/corr/abs-1907-10641}, AI2 Reasoning Challenge \citet{allenai:arc}, AI2 Reasoning Challenge 25-shots \citet{allenai:arc}, TruthfulQA \citet{lin2021truthfulqa}, massive multitask language understanding \citet{hendryckstest2021} and grade school math 8k \citet{gsm8k}. We report these results in \cref{tab:results_english} and plot the results in \cref{fig:eval_progression}.

\begin{table}[ht]
 \caption{Falcon2-11B evaluation results throughout the training. For each metric, we report the best accuracy and accuracy using normalized log probs. For reference, we also present the previous Falcon models evaluation scores. Here, HS-0, HS-10, Arc-E-0, WG-5, Arc-C-0, Arc-C-25, TQA-0, MMLU-5 and GSM8K-5 represent Hellaswag 0-shot, Hellaswag 10-shots, Arc Easy 0-shot, Winogrande 5-shots, Arc Challenge 0-shot, Arc Challenge 25-shots, TruthfulQA 0-shots, MMLU 5-shots and GSM8K 5-shots, respectively.}
  \centering
  \adjustbox{width=\textwidth}{
  \begin{tabular}{lcccccccccc}
    \toprule
    \textbf{Model} & \textbf{GT} & \textbf{HS-0} & \textbf{HS-10} & \textbf{ARC-E-0} & \textbf{WG-5} & \textbf{ARC-C-0} & \textbf{ARC-C-25} & \textbf{TQA-0} & \textbf{MMLU-5} & \textbf{GSM8K-5} \\
    \midrule
    Falcon 7B                   & 1500 & \num{76.31} & \num{78.1} & \num{74.74} & \num{67.17} & \num{43.43} & \num{47.86} & \num{34.3} & \num{35} & \num{4.62}\\
    Falcon 40B                  & 1000 & \textbf{82,82} & \textbf{85.28} & \textbf{81.86} & \num{76.4} & \textbf{54.69} & \textbf{61.86} & \num{41.65} & \num{56.89} & \num{21.46}\\
    \midrule
    Falcon2-11B \textit{stage 1} & 500  & \num{75.51} & \num{77.75} & \num{75.46} & \num{69.69} & \num{44.03} & \num{48.37} & \num{36.12} & \num{35.83} & -- \\
                                & 1000 & \num{76.51} & \num{79.19} & \num{76.81} &  --           & \num{45.39} & \num{52.64} & \num{40.74} & \num{45.56} & -- \\
                                & 1500 & \num{78.57} & \num{81.06} & \num{76.85} & \num{71.58} & \num{47.61} & \num{54.18} & \num{38.65} & \num{52.29} & -- \\
                                & 2000 & \num{78.66} & \num{81.23} & \num{75.92} & \num{72.21} & \num{46.92} & \num{52.55} & \num{38.92} & \num{51.88} & -- \\
                                & 2500 & \num{79.66} & \num{82.59} & \num{78.32} & \num{72.53} & \num{50.59} & \num{55.88} & \num{38.92} & \num{55.82} & -- \\
                                & 3000 & \num{79.92} & \num{83.34} & \num{79.16} & \num{73.79} & \num{51.71} & \num{59.47} & \num{41.31} & \num{57.43} & -- \\
                                & 3500 & \num{80.68} & \num{83.79} & \num{79.84} & \num{75.21} & \num{52.3}  & \num{59.47} & \num{42.01} & \num{58.42} & -- \\
                                & 4500 & \num{80.74} & \num{83.82} & \num{78.99} & \num{75.69} & \num{51.96} & \num{59.3}  & \num{40.89} & \num{59.08} & -- \\
    \midrule
    Falcon2-11B \textit{stage 2} & 4750 & \num{81.09} & \num{83.57} & \num{79.92} & \num{75.21} & \num{53.32} & \num{59.04} & \num{41.87} & \num{56.64} & -- \\
    \midrule
    Falcon2-11B \textit{stage 3} & 5000 & \num{81.15} & \num{83.69}  & \num{79.92} & \num{78.77}  & \num{53.32} & \num{60.15} & \num{41.31} & \textbf{59.63} & \num{22}\\
    \midrule
    Falcon2-11B \textit{stage 4} & 5500 & \num{82.07} & \num{82.91}  & \num{77.78} & \textbf{78.30}  & \num{50.17} & \num{59.73} & \textbf{52.56} & \num{58.37} & \textbf{53.83}\\
    \bottomrule
  \end{tabular}
}
  \label{tab:results_english}
\end{table}

\paragraph{Multilingual tasks} We evaluate multilingual capabilities following the selective tasks proposed on the Open Multilingual LLM Leaderboard \citep{lai2023openllmbenchmark} with results presented in \cref{tab:multilingual_leaderboard}, but also on the Belebele benchmark \citep{bandarkar2023belebele} and Wikilingua \citep{ladhak-wiki-2020}, for generative tasks, as presented in \cref{tab:multilingual_generative}. We note that the Open Multilingual LLM Leaderboard tasks are the translated versions of the English Open LLM Leaderboard tasks \cite{open-llm-leaderboard}.

\begin{table}[ht]
 \caption{HuggingFace open multilingual LLM evaluation leaderboard. In this table, we compare Falcon2-11B (ours) against Falcon-7B, Falcon-40B, Bloom-7B, Llama-7B, and Mistral-7B across different benchmarks. Evaluation scores on more languages are listed in \cref{sec:multi_eval}.}
  \centering
\begin{tabular}{llccccc}
\toprule
\textbf{Model}      & \textbf{Language} & \textbf{Arc-C-25}        & \textbf{Hellaswag}       & \textbf{MMLU-25}         & \textbf{TQA}             & \multicolumn{1}{l}{\textbf{Average}}   \\ \midrule
Mistral-7B & German (\textit{de}) & 41.23          & 58.73          & \textbf{40.46}          & 44.86 & 46.32                        \\
           & Spanish (\textit{es}) & 44.18          & 65.3           & \textbf{42.35} & 43.1  & 48.73                      \\
           & French (\textit{fr}) & 44.91          & 64.42          & \textbf{41.92} & 43.04          & 48.57                      \\
           & Italian (\textit{it}) & 43.19          & 60.88          & \textbf{39.71}          & 43.14          & 46.73                        \\
           & Dutch (\textit{nl}) & 40.03          & 57.94          & \textbf{41.42} & 43.28          & 45.66                      \\
           & Romanian (\textit{ro}) & 40.7           & 53.63          & \textbf{39.29}          & 43.63         & 44.31                      \\ \midrule
Falcon-7B  & French (\textit{fr}) & 37.29          & 64.08          & 28.37          & 34.03          & 40.94                      \\ \midrule
Llama-7B   & German (\textit{de}) & 35.1           & 49.9           & 29.9           & 38.3           & 38.3                         \\
           & Spanish (\textit{es}) & 36.8           & 56.4           & 30.3           & 37            & 40.12                       \\
           & French (\textit{fr}) & 37.3           & 55.7           & 30.5           & 39.9           & 40.85                        \\
           & Italian (\textit{it}) & 35.8           & 52            & 29.9           & 39.6           & 39.32                       \\
           & Dutch (\textit{nl}) & 33.6           & 48.7           & 29.8           & 40             & 38.02                      \\
           & Romanian (\textit{ro}) & 32.4           & 44.9           & 29.7           & 37            & 36             \\ \midrule
Bloom-7B   & German (\textit{de}) & 26.3           & 32.4           & 28.1           & 43.7           & 32.62                       \\
           & Spanish (\textit{es}) & 38.1           & 56.7           & 28.9           & 40.4           & 41.02                       \\
           & French (\textit{fr}) & 36.7           & 56.6           & 29.9           & 40.9           & 41.02                       \\
           & Italian (\textit{it}) & 29            & 40.8           & 27.6           & 43.7           & 35.27                       \\
           & Dutch (\textit{nl}) & 23.1           & 31.7           & 27.5           & 42.7           & 31.25                        \\
           & Romanian (\textit{ro}) & 26.9           & 31.8           & 27.4           & 46.1           & 33.05                        \\ \midrule
Falcon2-11B \textit{stage 4} & German (\textit{de}) & 43.7          & 67.96          &                38.3 & \textbf{47.53}           & \textbf{49.37}                     \\
           & Spanish (\textit{es}) & 46.2         & 73.63         & 37.9          & \textbf{46.43}          & \textbf{51.06}                       \\
           & French (\textit{fr}) & 45.8 & 72.41          & 39.53 & \textbf{47.30}          & \textbf{51.27}                      \\
           & Italian (\textit{it}) & 45.6          & \textbf{70.83} & 38.05 & \textbf{47.14}          & \textbf{50.42}                        \\
           & Dutch (\textit{nl}) & 41.7 & \textbf{69.05} & 38.29          & \textbf{48.81}          & \textbf{49.47}                        \\
           & Romanian (\textit{ro}) & 42.4          & \textbf{66.24} & 38.01 & \textbf{45.53}          & \textbf{48.04}                      \\ \midrule
Falcon-40B & German (\textit{de}) & \textbf{45.08} & \textbf{68.33} & 36.18          & 39.84          & 47.35  \\
           & Spanish (\textit{es}) & \textbf{48.46} & \textbf{73.94} & 37.2           & 38.96          & 49.64    \\
           & French (\textit{fr}) & \textbf{47.64}          & \textbf{72.86} & 37.3           & 38.49          & 49.07  \\
           & Italian (\textit{it}) & \textbf{46.27} & 70.21          & 36.36          & 40.67          & 48.37 \\
           & Dutch (\textit{nl}) & \textbf{42.85}          & 68.39          & 36.48          & 40.85          & 47.14  \\
           & Romanian (\textit{ro}) & \textbf{43.18} & 66          & 35.67          & 39.81          & 46.2 \\ \bottomrule
\end{tabular}
  \label{tab:multilingual_leaderboard}
\end{table}

\begin{table}[ht]
 \caption{Detailed performance on Belebele and Wikilingua tasks for Mistral 7B and Falcon2-11B, \textit{stages 3} and \textit{4}. Note that Belebele is not available in Dutch, and Wikilingua is not available in Polish, Romanian, Danish and Swedish.}
  \centering
  \begin{tabular}{llcccccc}
    \toprule
    \textbf{Model} & \textbf{Language} & \multicolumn{2}{c}{\textbf{Belebele}} & \multicolumn{4}{c}{\textbf{Wikilingua}} \\
    \cmidrule(r){3-4} \cmidrule(r){5-8}
     &  & 0 shot & 5 shots & Rouge1 & Rouge2 & RougeL & RougeGeo \\
    \midrule
    Mistral-7B & English (\textit{en}) & 32.48 & \textbf{79.42} & \textbf{0.2770} & 0.0883 & 0.2126 & 0.1732 \\
               & German (\textit{de}) & 21.69 & \textbf{66.55} & 0.1530 & 0.0340 & 0.1137 & 0.0839 \\
               & Spanish (\textit{es}) & 26.47 & \textbf{68.00} & 0.2319 & 0.0729 & 0.1674 & 0.1414 \\
               & French (\textit{fr}) & 25.58 & \textbf{67.45} & 0.1914 & 0.0571 & 0.1420 & 0.1158 \\
               & Italian (\textit{it}) & 22.25 & \textbf{65.88} & 0.1784 & 0.0459 & 0.1329 & 0.1029 \\
               & Portuguese (\textit{pt}) & 23.25 & \textbf{67.56} & 0.2278 & 0.0618 & 0.1561 & 0.1300 \\
               & Czech (\textit{cz}) & 31.81 & \textbf{64.32} & 0.1802 & 0.0402 & 0.1233 & 0.0963 \\
               & Dutch (\textit{nl}) & -- & -- & 0.2055 & 0.0474 & 0.1527 & 0.1142 \\
               & Polish (\textit{pl}) & 29.92 & \textbf{62.53} & -- & -- & -- & -- \\
               & Romanian (\textit{ro}) & 26.03 & 63.98 & -- & -- & -- & -- \\
               & Danish (\textit{da}) & 26.03 & \textbf{61.52} & -- & -- & -- & -- \\
               & Swedish (\textit{sv}) & 28.03 & \textbf{66.33} & -- & -- & -- & -- \\

    \midrule
    Falcon-11B \textit{stage 3} & English (\textit{en}) & 40.93 & 67.56 & 0.2741 & \textbf{0.0922} & \textbf{0.2187} & \textbf{0.1768} \\
               & German (\textit{de}) & 34.82 & 63.43 & \textbf{0.2125} & \textbf{0.0607} & \textbf{0.1568} & \textbf{0.1265} \\
               & Spanish (\textit{es}) & 34.37 & 59.40 & \textbf{0.2805} & \textbf{0.1079} & \textbf{0.2070} & \textbf{0.1843} \\
               & French (\textit{fr}) & 31.92 & 60.07 & \textbf{0.2660} & \textbf{0.1006} & \textbf{0.2004} & \textbf{0.1750} \\
               & Italian (\textit{it}) & 32.81 & 60.63 & \textbf{0.2331} & \textbf{0.0754} & \textbf{0.1777} & \textbf{0.1462} \\
               & Portuguese (\textit{pt}) & 38.15 & 59.61 & \textbf{0.2613} & \textbf{0.0843} & \textbf{0.1831} & \textbf{0.1592} \\
               & Czech (\textit{cz}) & 36.92 & 54.25 & \textbf{0.2174} & \textbf{0.0629} & \textbf{0.1568} & \textbf{0.1290} \\
               & Dutch (\textit{nl}) & -- & -- & \textbf{0.2680} & \textbf{0.0817} & \textbf{0.2046} & \textbf{0.1648} \\
               & Polish (\textit{pl}) & 30.81 & 55.37 & -- & -- & -- & -- \\
               & Romanian (\textit{ro}) & 30.70 & 62.30 & -- & -- & -- & -- \\
               & Danish (\textit{da}) & 30.92 & 51.68 & -- & -- & -- & -- \\
               & Swedish (\textit{sv}) & 38.82 & 59.17 & -- & -- & -- & -- \\
    \midrule
    Falcon-11B \textit{stage 4} & English (\textit{en}) & \textbf{66.41} & 73.27 & 0.2658 & 0.0773 & 0.1908 & 0.1577 \\
               & German (\textit{de}) & \textbf{64.63} & 62.30 & 0.1309 & 0.0343 & 0.0951 & 0.0753 \\
               & Spanish (\textit{es}) & \textbf{66.18} & 67.45 & 0.1518 & 0.0507 & 0.1092 & 0.0944 \\
               & French (\textit{fr}) & \textbf{60.62} & 66.22 & 0.1416 & 0.0482 & 0.1040 & 0.0892 \\
               & Italian (\textit{it}) & \textbf{60.40} & 63.65 & 0.1493 & 0.0416 & 0.1077 & 0.0875 \\
               & Portuguese (\textit{pt}) & \textbf{61.18} & 63.65 & 0.1780 & 0.0522 & 0.1247 & 0.1050 \\
               & Czech (\textit{cz}) & \textbf{62.07} & 62.08 & 0.1254 & 0.0326 & 0.0868 & 0.07077 \\
               & Dutch (\textit{nl}) & -- & -- & 0.1487 & 0.0371 & 0.1073 & 0.0839 \\
               & Polish (\textit{pl}) & \textbf{59.73} & 59.62 & -- & -- & -- & -- \\
               & Romanian (\textit{ro}) & \textbf{64.07} & \textbf{66.33} & -- & -- & -- & -- \\
               & Danish (\textit{da}) & \textbf{56.73} & 57.38 & -- & -- & -- & -- \\
               & Swedish (\textit{sv}) & \textbf{66.74} & \textbf{66.33} & -- & -- & -- & -- \\

    \bottomrule
  \end{tabular}
  \label{tab:multilingual_generative}
\end{table}

When analyzing the multilingual evaluation results in \cref{tab:multilingual_generative}, we first observe that both Falcon2-11B \textit{stage 3} and \textit{stage 4} perform better on Belebele 0-shot evaluation in comparison to Mistral-7B, however, Mistral-7B scores tend to be better in the 5-shot scenario. Secondly, a comparison between the results on selective and generative evaluation tasks shows that Mistral-7B has good performance on selective tasks in all languages presented in \cref{tab:multilingual_leaderboard} but worse performance on generative tasks in languages other than English (see \cref{tab:multilingual_generative}).  This is interesting to note, as Mistral-7B does not claim to have any multilingual capabilities. Similarly, the Falcon2-11B \textit{stage 4} shows increased performance on selective tasks both in English (see \cref{tab:results_english}) and other languages (as shown in \cref{tab:multilingual_leaderboard}), but degrades on generative tasks in the multilingual setting, particularly when compared to Falcon2-11B \textit{stage 3}. This can be explained as the data mixture used in \textit{stage 4} was particularly focused on English data. This hints at the possibility that a model with strong English performance can also achieve high scores on selective multilingual tasks relying on inner translation capabilities that can be achieved by adding some multilingual content to its training data. However, generative tasks show the model's intrinsic performance in a given language.

\paragraph{Code generation}
We report in \cref{tab:HumanEval} the evaluation of the Falcon family of models on Python code generation. Their capabilities are evaluated through pass@1 accuracy on the HumanEval (0-shot) benchmark \citep{chen2021evaluating}. The temperature is set to $0.2$ and the decoding strategy is nucleus sampling with $\text{top-p}=0.95$. The maximum output length is $512$ and 50 completions are produced for each prompt\footnote{Settings are used for code generation from \url{https://huggingface.co/spaces/bigcode/bigcode-models-leaderboard}}.

\begin{table}[ht]
\caption{Pass@1 metric on the HumanEval code generation benchmark evaluated on the Falcon and Falcon2 series.}
    \centering
    \begin{tabular}{lc}
    \toprule
        \textbf{Model} & \textbf{pass@1} \\
        \midrule
        Falcon-7B &  10\\
        Falcon-40B & 23.1\\
        Falcon-180B & 35.3\\
        \midrule
        Falcon2-11B \textit{stage 1} & 15\\
        Falcon2-11B \textit{stage 2} & 18.1 \\
        Falcon2-11B \textit{stage 3} & 17.9\\
        Falcon2-11B \textit{stage 4} & 29.6\\
        \bottomrule
    \end{tabular}
    \label{tab:HumanEval}
\end{table}

\section{Multimodal Extension: Falcon2-11B VLM\label{falcon_vlm}}\label{sec:vlm}
In this section, we extend the Falcon2-11B to a vision-language model (VLM) for additionally handling image inputs and answering the queries corresponding to the images. To achieve this, we integrate a vision encoder with our Falcon2-11B and train with image-text data.  

\noindent\textbf{Architecture}
The overall architecture of our Falcon2-11B VLM is as follows. A pre-trained CLIP ViT-L/14~\citep{radford2021learning} is employed as the visual encoder $f_e(\cdot)$ for encoding the input images $\bm{X}$ into local patch (visual) features $\bm{H}_e$. These patch features are further transformed to the LLM input word embedding space through a two-layer multimodal projector $f_{\mathrm{proj}}(\cdot)$. The projected patch embeddings $\bm{H}_\mathrm{p}$ act as soft-token inputs to the Falcon LLM $f_{\mathrm{LLM}}(\cdot)$. The language instruction $\bm{X}_t$ corresponding to the image $\bm{X}$ is tokenized and transformed into $d$-dimensional embeddings $\bm{H}_t$. Next, the patch embeddings $\bm{H}_p$ and instruction embeddings $\bm{H}_t$ are concatenated along the sequence length and provided as input to the LLM $f_{\mathrm{LLM}}(\cdot)$. The LLM then outputs an appropriate response $\bm{X}_r$. We use the Falcon2-11B-chat model as the LLM for this multimodal extension. 
Furthermore, similar to LLaVA-NeXT \citet{liu2024llavanext}, for enhancing the VLM's perception of fine-grained details w.r.t small objects in images, we employ a dynamic encoding mechanism at high-resolution for image inputs. This also mitigates the model hallucination that might occur when presented with low-resolution images. 

\subsection{Training}
The training of the Falcon VLM is done in two stages, \emph{viz.}, pretraining and finetuning. In both stages, the visual encoder weights are kept frozen. In the pretraining stage, the LLM is kept frozen and only the multimodal projector is trained on 558K image-caption pairs from the LLaVA dataset \citet{liu2023improvedllava}. This enables the multimodal projector to learn a mapping from visual to text embedding space. During finetuning, both the projector and LLM weights are trained on a corpus of 1.2M image-text instruction data, which also includes multi-round conversations. The finetuning data is a combination of conversational data from LLaVA-665K \citet{liu2023improvedllava}, ShareGPT4V \citet{chen2023sharegpt4v} and MMInstruct \citet{li2023m3it} covering AI2D \citet{kembhavi2016diagram}, CLEVR \citet{johnson2017clevr}, Multi30k \citet{multi30k}, NLVR \citet{nlvr}, OCR-VQA \citet{ocrvqa}, STVQA \citet{stvqa}, VisualDialog \citet{visdial}, Visual-MRC \citet{VisualMRC2021}, ViQuAE \citet{lerner2022viquae}, and VCR \citet{zellers2019vcr} datasets.

\subsection{Evaluation}
\cref{tab:vlm_results} shows the performance comparison between the Falcon2-11B VLM and publicly available VLMs (of similar size) on fourteen popular challenging benchmarks: MME \citet{fu2023mme}, GQA \citet{hudson2019gqa}, ScienceQA \citet{scienceqa}, POPE \citet{pope}, VQAv2 \citet{vqav2}, TextVQA \citet{textvqa}, MM-Bench \citet{MMBench}, SEED-IMG \citet{seedbench}, AI2D \citet{kembhavi2016diagram}, DocVQA \citet{mathew2021docvqa}, InfoVQA \citet{mathew2022infographicvqa}, MathVista \citet{lu2023mathvista}, HallusionBench \citet{guan2023hallusionbench}, and RealWorldQA \citet{realworldqa}.
Our Falcon VLM enhances the visual-language reasoning capability, achieving favorable performance overall (Average) across the benchmarks.

\begin{table}[ht]
\caption{VLM benchmark performance comparison. In comparison to LLaVA-NeXT \citet{liu2024llavanext} (LlaVA1.6) variants, our Falcon VLM achieves favorable performance overall (Average) across the fourteen challenging benchmarks. Here, VQA${}^T$, VQA${}^D$, VQA${}^{I}$, SEED${}^I$, MMB, HB, and QA${}^{RW}$ denote TextVQA, DocVQA, InfoVQA, Seed-Bench2 (Image), MM-Bench, HallusionBench, and RealWorldQA, respectively. For computing the average (Avg.), the total score of MME is divided by 20 and added to the remaining benchmark results.}
\label{tab:vlm_results}
\small
\centering
\adjustbox{width=\columnwidth}{
\begin{tabular}{lccccccccccccccc}
\toprule
\textbf{Model} & \textbf{MME} & \textbf{GQA} & \textbf{SQA} & \textbf{POPE} & \textbf{VQA${}^{v2}$} & \textbf{VQA${}^T$} & \textbf{MMB} & \textbf{SEED${}^I$} & 
\textbf{AI2D} & \textbf{VQA${}^{D}$} & \textbf{VQA${}^{I}$} & \textbf{MV} & \textbf{HB} & \textbf{QA${}^{RW}$} & \textbf{Avg.} \\
\midrule
Falcon2-11B VLM & \textbf{1589/343} & 64.5 & \textbf{74.9} & \textbf{88.4} & 82.1 & 66.7 & \textbf{72.0} & \textbf{72.3} & \textbf{78.3} & 76.9 & 42.6 & 37.2 & \textbf{48.7} & \textbf{61.7} & \textbf{68.8} \\
LLaVA1.6-Vicuna-7B & 1519/332 & 64.2 & 70.1 & 86.5 & 81.8 & 64.9 & 67.4 & 70.2 & 66.6 & 74.4 & 37.1 & 34.4 & 41.5 & -- & 65.5 \\
LLaVA1.6-Vicuna-13B & 1575/326 & \textbf{65.4} & 73.6 & 86.2 & \textbf{82.8} & \textbf{67.1} & 70.0 & 71.9 & 70.0 & \textbf{77.4} & 41.3 & 35.1 & 44.5 & -- & 67.7 \\
LLaVA1.6-Mistral-7B & 1498/321 & 64.8 & 72.8 & 86.7 & 82.2 & 65.7 & 68.7 & 72.2 & 60.8 & 72.2 & \textbf{43.8} & \textbf{37.4} & 41.7 & 54.4 & 65.3 \\
\bottomrule
\end{tabular}
}
\end{table}

\section{Model Availability and License}\label{sec:license}
The Falcon 2 models are made available under the Falcon 2 11B TII License \footnote{\url{https://falconllm-staging.tii.ae/falcon-2-terms-and-conditions.html}}, a permissive Apache 2.0-based software license which includes an acceptable use policy \footnote{\url{https://falconllm.tii.ae/falcon-2-acceptable-use-policy.html}} that promotes the responsible use of AI. The models can be downloaded on the HuggingFace hub using the following links:
\begin{itemize}[topsep=0pt,itemsep=0pt,parsep=5pt,partopsep=0pt]
    \item Falcon2-11B (LLM): \url{https://huggingface.co/tiiuae/falcon-11B};
    \item Falcon2-11B-VLM: \url{https://huggingface.co/tiiuae/falcon-11B-vlm}.
\end{itemize}

\section*{Acknowledgments}
The authors are thankful to the HuggingFace leaderboard team for support in promptly evaluating our models on the Open LLM Leaderboard.

\newpage
\bibliography{main}
\bibliographystyle{apalike}
\newpage

\appendix

\section{Language-tuned quality filters}
\label{sec:filter_details}
To further increase the quality of the multilingual training samples obtained from CommonCrawl data, we fully extend the pre-processing data pipeline used for English samples, as detailed in \citet{refinedweb}, to ten other languages. Specifically, we add language-tuned line-wise quality filters, and two heuristics-based filters inspired from \citet{gopher}, using the values presented in \cref{tab:lang_tuned_values}.

Based on word frequency versus word length (in letters) distributions of naturally occurring text in each language, Romance languages (French, Italian, Portuguese, Romanian, Spanish) tend to have slightly narrower distributions compared to Czech, Dutch, German, Polish, and Swedish. Stop words are selected using the most frequently occurring word lists for each language, from which we exclude nouns, pronouns, adjectives, adverbs, and most verb conjugations, and focus on articles, prepositions, conjunctions, and some common verb forms. As in \citet{gopher}, we set the minimum passing threshold to two stop words per sample.

\begin{table}[h!]
\caption{Values used for language-tuned versions of average word length and minimum stop word filters from \citet{gopher}.}
  \centering
  \begin{tabular}{lcl}
    \toprule
    \textbf{Language} & \textbf{Character per word range} & \textbf{Stop words} \\
    \midrule
    Czech (\textit{cs}) & 2--13 & a, k, ke, z, ze, u, to, do, mít, s, se, na, v, ve, je, jsem \\
    German (\textit{de}) & 3--13 & das, sein, zu, von, und, haben, mit \\
    Spanish (\textit{es}) & 3--11 & el, la, los, las, en, a, de, del, y, con, que, es, ha \\
    French (\textit{fr}) & 3--11 & les, dans, un, une, de, et, ou, avec, cela, c’est, à, comme, que \\
    Italian (\textit{it}) & 3--11 & il, in, a, da, di, che, con, per, sono, è, era, io, lui \\
    Dutch (\textit{nl}) & 3--13 & de, zijn, naar, van, en, dat, hebben, met \\
    Polish (\textit{pl}) & 2--13 & do, że, i, co, to, mieć, z, w, ze, na, jestem, jest \\
    Portuguese (\textit{pt}) & 3--11 & o, em, a, de, e, com, que, é, para \\
    Romanian (\textit{ro}) & 3--11 & o, un, care, este, către, spre, din, în, și, sau, să, ca, cu, la, de \\
    Swedish (\textit{sv}) & 3--13 & det, vara, till, av, och, har, med \\
    \bottomrule
  \end{tabular}
  \label{tab:lang_tuned_values}
\end{table}

\section{Programming languages}
\label{sec:code_languages}
The code slice of the pre-training data includes samples in 43 languages, which are, in alphabetical order: Assembly, Batchfile, C, CMake, C++, C\#, CSS, Dart, Dockerfile, Fortran, Go, Haskell, HTML, Java, JavaScript, Julia, Kotlin, Labview, Lua, Makefile, Maple, Markdown, Mathematica, Matlab, Nix, Objective-C++, Octave, Perl, PHP, Powershell, Python, R, Ruby, Rust, SAS, Scala, Scilab, Shell, SQL, Swift, TeX, TypeScript, Visual Basic.

\newpage

\section{Evaluation Scores on other languages}
\label{sec:multi_eval}

\begin{table}[ht]
 \caption{Evaluations on Open Multilingual LLM Leaderboard. The languages are sorted by language ID, from the list of ISO 639 language codes \protect\footnotemark.}
  \centering
\begin{tabular}{llcccc}
\toprule
\textbf{Model}      & \textbf{Language} & \textbf{Arc-C-25}        & \textbf{Hellaswag-0}       & \textbf{MMLU-25}         & \textbf{TQA-0}     \\ \midrule
Falcon2-11B \textit{stage 4} & Arabic (\textit{ar}) &   25.32        & 32.42        & 28.04          & 49.66                         \\
           & Bengali (\textit{bn}) & 24.37          & 28.57           & 27.88 & 51.65                     \\
           & Catalan (\textit{ca}) & 38.33          & 56.80          & 34.59 & 47.33                    \\
           & Danish (\textit{da}) & 34.79          & 57.01          & 35.68          & 49.31                            \\
           & Basque (\textit{eu}) & 25.31          & 29.22          & 28.47 & 42.31                              \\
           & Gujarati (\textit{gu}) & 24.13          & 28.71          & 27.95          & 42.61                          \\ 
           & Hindi (\textit{hi}) & 25.17          & 29.49          & 28.59          & 48.87                  \\ 
           & Croatian (\textit{hr}) & 29.85          & 45.06          & 31.32          & 48.32                               \\
           & Hungarian (\textit{hu}) & 24.74        & 31.79          & 29.02          & 46.71                             \\ 
           & Armenian (\textit{hy}) & 25.72          & 27.95          & 26.80          & 45.84                           \\ 
           & Indonesian (\textit{id}) & 34.18          & 51.35          & 33.25          & 46.96                             \\
           & Kannada (\textit{kn}) & 24.56          & 29.67          & 27.97          & 47.29                           \\
           & Malayalam (\textit{ml}) & 27.49          & 29.32          & 27.14          & 49.66                             \\
           & Marathi (\textit{mr}) & 24.84          & 29.03          & 28.12          & 49.41                  \\
           & Nepali (\textit{ne}) & 23.69          & 29.12          & 28.32          & 44.23                          \\
           & Russian (\textit{ru}) & 34.13          & 51.35          & 32.01          & 47.59                          \\
           & Slovak (\textit{sk}) & 36.27          & 55.01          & 34.55          & 46.41                        \\
           & Serbian (\textit{sr}) & 31.13          & 44.79          & 31.54          & 50.16                              \\
           & Tamil (\textit{ta}) & 28.19          & 28.45          & --          & 49.35                           \\
           & Telugu (\textit{te}) & 25.43          & 29.61          & 27.56          & 48.25                          \\
           & Ukrainian (\textit{uk}) & 31.05          & 41.14          & 29.43          & 46.70                             \\
           & Vietnamese (\textit{vi}) & 27.09          & 38.62          & 29.93          & 44.31                             \\
           & Chinese (\textit{zh}) & 39.82          & 59.09          & 35.41         & 47.13                            \\
           
           \bottomrule
\end{tabular}
  \label{tab:multilingual_leaderboard_full}
\end{table}

\footnotetext{\url{https://en.wikipedia.org/wiki/List_of_ISO_639_language_codes}}

\newpage

\section{VLM Architecture Overview}
\input{figure_vlm.tex}

\end{document}

%% file: conversation_graph.tex
\GRAPH convosGraph

\begin{figure}[H]
\begin{center}

\begin{tikzpicture}[
dot/.style = {circle, fill, minimum size=#1,
                inner sep=0pt, outer sep=0pt},
               scale=0.45, every node/.style={scale=0.45}
]]
\draw[fill=fine-tuning] (0.0, -0.7) rectangle (1.4, 0.7);
\node at (0.7, 0.0) {\large  \phantom{lj}Do\phantom{lj}};
\draw[fill=fine-tuning] (1.4, -0.7) rectangle (2.8, 0.7);
\node at (2.0999999999999996, 0.0) {\large  \phantom{lj}you\phantom{lj}};
\draw[fill=fine-tuning] (2.8000000000000003, -0.7) rectangle (4.2, 0.7);
\node at (3.5, 0.0) {\large  \phantom{lj}like\phantom{lj}};
\draw[fill=fine-tuning] (4.200000000000001, -0.7) rectangle (5.6000000000000005, 0.7);
\node at (4.9, 0.0) {\large  \phantom{lj}Venice?\phantom{lj}};
\draw[fill=objective] (6.4, 0.10000000000000009) rectangle (7.800000000000001, 1.5);
\node at (7.1000000000000005, 0.8) {\large  \phantom{lj}Not\phantom{lj}};
\draw[fill=objective] (7.800000000000001, 0.10000000000000009) rectangle (9.200000000000001, 1.5);
\node at (8.5, 0.8) {\large  \phantom{lj}really,\phantom{lj}};
\draw[fill=objective] (9.200000000000003, 0.10000000000000009) rectangle (10.600000000000003, 1.5);
\node at (9.900000000000002, 0.8) {\large  \phantom{lj}it's\phantom{lj}};
\draw[fill=objective] (10.600000000000005, 0.10000000000000009) rectangle (12.000000000000005, 1.5);
\node at (11.300000000000004, 0.8) {\large  \phantom{lj}always\phantom{lj}};
\draw[fill=objective] (12.000000000000007, 0.10000000000000009) rectangle (13.550000000000007, 1.5);
\node at (12.750000000000006, 0.8) {\large  \phantom{lj}flooded\phantom{lj}};
\draw[fill=architecture] (14.200000000000008, 0.9000000000000001) rectangle (15.600000000000009, 2.3);
\node at (14.90000000000001, 1.6) {\large  \phantom{lj}Oh,\phantom{lj}};
\draw[fill=architecture] (15.600000000000007, 0.9000000000000001) rectangle (17.000000000000007, 2.3);
\node at (16.300000000000008, 1.6) {\large  \phantom{lj}I\phantom{lj}};
\draw[fill=architecture] (17.000000000000007, 0.9000000000000001) rectangle (18.400000000000006, 2.3);
\node at (17.700000000000006, 1.6) {\large  \phantom{lj}agree\phantom{lj}};
\draw[fill=evaluation] (14.200000000000008, -0.7) rectangle (15.600000000000009, 0.7);
\node at (14.90000000000001, 0.0) {\large  \phantom{lj}Really?\phantom{lj}};
\draw[fill=evaluation] (15.600000000000007, -0.7) rectangle (17.000000000000007, 0.7);
\node at (16.300000000000008, 0.0) {\large  \phantom{lj}it's\phantom{lj}};
\draw[fill=evaluation] (17.000000000000007, -0.7) rectangle (18.400000000000006, 0.7);
\node at (17.700000000000006, 0.0) {\large  \phantom{lj}so\phantom{lj}};
\draw[fill=evaluation] (18.400000000000006, -0.7) rectangle (19.800000000000004, 0.7);
\node at (19.100000000000005, 0.0) {\large  \phantom{lj}nice!\phantom{lj}};
\draw[fill=adaptation] (6.4, -1.5) rectangle (7.800000000000001, -0.10000000000000009);
\node at (7.1000000000000005, -0.8) {\large  \phantom{lj}Yes!\phantom{lj}};
\draw[fill=adaptation] (7.800000000000001, -1.5) rectangle (9.200000000000001, -0.10000000000000009);
\node at (8.5, -0.8) {\large  \phantom{lj}It's\phantom{lj}};
\draw[fill=adaptation] (9.200000000000003, -1.5) rectangle (10.600000000000003, -0.10000000000000009);
\node at (9.900000000000002, -0.8) {\large  \phantom{lj}unique!\phantom{lj}};

\draw[fill=fine-tuning] (0.0, -4.7) rectangle (1.4, -3.3000000000000003);
\node at (0.7, -4.0) {\large  \phantom{lj}Do\phantom{lj}};
\draw[fill=fine-tuning] (1.4, -4.7) rectangle (2.8, -3.3000000000000003);
\node at (2.0999999999999996, -4.0) {\large  \phantom{lj}you\phantom{lj}};
\draw[fill=fine-tuning] (2.8000000000000003, -4.7) rectangle (4.2, -3.3000000000000003);
\node at (3.5, -4.0) {\large  \phantom{lj}like\phantom{lj}};
\draw[fill=fine-tuning] (4.200000000000001, -4.7) rectangle (5.6000000000000005, -3.3000000000000003);
\node at (4.9, -4.0) {\large  \phantom{lj}Venice?\phantom{lj}};
\draw[fill=objective] (5.6000000000000005, -4.7) rectangle (7.0, -3.3000000000000003);
\node at (6.300000000000001, -4.0) {\large  \phantom{lj}Not\phantom{lj}};
\draw[fill=objective] (6.999999999999999, -4.7) rectangle (8.399999999999999, -3.3000000000000003);
\node at (7.699999999999999, -4.0) {\large  \phantom{lj}really,\phantom{lj}};
\draw[fill=objective] (8.399999999999999, -4.7) rectangle (9.799999999999997, -3.3000000000000003);
\node at (9.099999999999998, -4.0) {\large  \phantom{lj}it's\phantom{lj}};
\draw[fill=objective] (9.799999999999997, -4.7) rectangle (11.199999999999996, -3.3000000000000003);
\node at (10.499999999999996, -4.0) {\large  \phantom{lj}always\phantom{lj}};
\draw[fill=objective] (11.199999999999996, -4.7) rectangle (12.749999999999994, -3.3000000000000003);
\node at (11.949999999999995, -4.0) {\large  \phantom{lj}flooded\phantom{lj}};
\draw[fill=architecture] (12.749999999999994, -4.7) rectangle (13.999999999999995, -3.3000000000000003);
\node at (13.299999999999994, -4.0) {\large  \phantom{lj}Oh,\phantom{lj}};
\draw[fill=architecture] (13.999999999999996, -4.7) rectangle (15.399999999999997, -3.3000000000000003);
\node at (14.699999999999996, -4.0) {\large  \phantom{lj}I\phantom{lj}};
\draw[fill=architecture] (15.399999999999999, -4.7) rectangle (16.799999999999997, -3.3000000000000003);
\node at (16.099999999999998, -4.0) {\large  \phantom{lj}agree\phantom{lj}};

\draw[pattern=north west lines, pattern color=fine-tuning] (0.0, -6.7) rectangle (1.4, -5.3000000000000003);
\node at (0.7, -6.0) {\large  \phantom{lj}Do\phantom{lj}};
\draw[pattern=north west lines, pattern color=fine-tuning] (1.4, -6.7) rectangle (2.8, -5.3000000000000003);
\node at (2.0999999999999996, -6.0) {\large  \phantom{lj}you\phantom{lj}};
\draw[pattern=north west lines, pattern color=fine-tuning] (2.8000000000000003, -6.7) rectangle (4.2, -5.3000000000000003);
\node at (3.5, -6.0) {\large  \phantom{lj}like\phantom{lj}};
\draw[pattern=north west lines, pattern color=fine-tuning] (4.200000000000001, -6.7) rectangle (5.6000000000000005, -5.3000000000000003);
\node at (4.9, -6.0) {\large  \phantom{lj}Venice?\phantom{lj}};
\draw[pattern=north west lines, pattern color=objective] (5.6000000000000005, -6.7) rectangle (7.0, -5.3000000000000003);
\node at (6.300000000000001, -6.0) {\large  \phantom{lj}Not\phantom{lj}};
\draw[pattern=north west lines, pattern color=objective] (6.999999999999999, -6.7) rectangle (8.399999999999999, -5.3000000000000003);
\node at (7.699999999999999, -6.0) {\large  \phantom{lj}really,\phantom{lj}};
\draw[pattern=north west lines, pattern color=objective] (8.399999999999999, -6.7) rectangle (9.799999999999997, -5.3000000000000003);
\node at (9.099999999999998, -6.0) {\large  \phantom{lj}it's\phantom{lj}};
\draw[pattern=north west lines, pattern color=objective] (9.799999999999997, -6.7) rectangle (11.199999999999996, -5.3000000000000003);
\node at (10.499999999999996, -6.0) {\large  \phantom{lj}always\phantom{lj}};
\draw[pattern=north west lines, pattern color=objective] (11.199999999999996, -6.7) rectangle (12.749999999999994, -5.3000000000000003);
\node at (11.949999999999995, -6.0) {\large  \phantom{lj}flooded\phantom{lj}};
\draw[fill=evaluation] (12.749999999999994, -6.7) rectangle (14.099999999999995, -5.3000000000000003);
\node at (13.399999999999994, -6.0) {\large  \phantom{lj}Really?\phantom{lj}};
\draw[fill=evaluation] (14.099999999999995, -6.7) rectangle (15.399999999999997, -5.3000000000000003);
\node at (14.699999999999996, -6.0) {\large  \phantom{lj}it's\phantom{lj}};
\draw[fill=evaluation] (15.399999999999999, -6.7) rectangle (16.799999999999997, -5.3000000000000003);
\node at (16.099999999999998, -6.0) {\large  \phantom{lj}so\phantom{lj}};
\draw[fill=evaluation] (16.799999999999997, -6.7) rectangle (18.199999999999996, -5.3000000000000003);
\node at (17.499999999999996, -6.0) {\large  \phantom{lj}nice!\phantom{lj}};


\draw[pattern=north west lines, pattern color=fine-tuning] (0.0, -8.7) rectangle (1.4, -7.3000000000000003);
\node at (0.7, -8.0) {\large  \phantom{lj}Do\phantom{lj}};
\draw[pattern=north west lines, pattern color=fine-tuning] (1.4, -8.7) rectangle (2.8, -7.3000000000000003);
\node at (2.0999999999999996, -8.0) {\large  \phantom{lj}you\phantom{lj}};
\draw[pattern=north west lines, pattern color=fine-tuning] (2.8000000000000003, -8.7) rectangle (4.2, -7.3000000000000003);
\node at (3.5, -8.0) {\large  \phantom{lj}like\phantom{lj}};
\draw[pattern=north west lines, pattern color=fine-tuning] (4.200000000000001, -8.7) rectangle (5.6000000000000005, -7.3000000000000003);
\node at (4.9, -8.0) {\large  \phantom{lj}Venice?\phantom{lj}};
\draw[fill=adaptation] (5.6000000000000005, -8.7) rectangle (7.0, -7.3000000000000003);
\node at (6.300000000000001, -8.0) {\large  \phantom{lj}Yes!\phantom{lj}};
\draw[fill=adaptation] (6.999999999999999, -8.7) rectangle (8.399999999999999, -7.3000000000000003);
\node at (7.699999999999999, -8.0) {\large  \phantom{lj}It's\phantom{lj}};
\draw[fill=adaptation] (8.399999999999999, -8.7) rectangle (9.799999999999997, -7.3000000000000003);
\node at (9.099999999999998, -8.0) {\large  \phantom{lj}unique!\phantom{lj}};

\draw[pattern=north west lines, pattern color=black] (19.049999999999994, -4.25) rectangle (19.549999999999994, -3.75);
\node[right] at (19.549999999999994, -4) {\Large Loss masked};



\draw[->,] (5.6000000000000005, 0.0) to [bend left = 0] (6.4, 0.8);
\draw[->,] (13.550000000000007, 0.8) to [bend left = 0] (14.200000000000008, 1.6);
\draw[->,] (13.550000000000007, 0.8) to [bend left = 0] (14.200000000000008, 0.0);
\draw[->,] (5.6000000000000005, 0.0) to [bend left = 0] (6.4, -0.8);

\end{tikzpicture}
\captionof{figure}{Visualization of conversation tree flattening and loss masking. In this simple example, the token count of the conversation tree is 19, while the flattened conversation tree contains the 19 initial tokens, but also contains 13 repeated tokens with masked loss, producing an overhead of $+68\%$.} 
\label{fig:convosGraph}

\end{center}
\end{figure}

\ENDGRAPH

%% file: figure_vlm.tex
\begin{figure}[H]
\begin{center}

\resizebox{\textwidth}{!}{
\begin{tikzpicture}[line width=2]
    \def\gridWidth{14}
    \def\gridCenter{8}
    \def\gridHeight{6}
    \def\rectWidth{0.8}
    \def\rectHeight{0.8}
    \def\spacing{0.2}
    \def\halfImages{3}
    \pgfmathsetmacro{\thresholdImages}{2*\halfImages + 1}
    \pgfmathsetmacro{\tokenDots}{\halfImages + 1}
    \def\braceHeight{1}
    \def\arrowHeight{0.8}
    \def\imageSide{5}
    \def\patchSide{0.8}
    \def\patchSidecm{0.8cm}
    \def\nGeneratedTokens{7}
    \pgfmathsetmacro{\xDots}{\tokenDots + 0.5*\rectWidth}
    \pgfmathsetmacro{\yDots}{0.5*\rectHeight}
    \pgfmathsetmacro{\yFalcon}{\spacing + \rectHeight}
    \pgfmathsetmacro{\falconWidth}{\gridWidth - \spacing + \nGeneratedTokens - 1}
    \pgfmathsetmacro{\totalTokens}{\gridWidth + \nGeneratedTokens - 1}
    \pgfmathsetmacro{\yProjector}{-\rectHeight - \arrowHeight} 
    \pgfmathsetmacro{\yTokens}{\yProjector - \rectHeight - \spacing}
    \pgfmathsetmacro{\yVisionEncoder}{\yProjector - \arrowHeight - \rectHeight - \rectHeight - \spacing}
    \pgfmathsetmacro{\projectorWidth}{\thresholdImages-\spacing}
    \pgfmathsetmacro{\tokenizerWidth}{\gridWidth - \thresholdImages - \spacing}
    \pgfmathsetmacro{\embWidth}{\tokenizerWidth + \nGeneratedTokens - 1}

    \pgfmathsetmacro{\xTokenizer}{\thresholdImages-\spacing}
    \pgfmathsetmacro{\yPatches}{\yVisionEncoder-\spacing-0.5*\patchSide}
    \pgfmathsetmacro{\yImage}{\yPatches - 0.5*\imageSide -0.5*\patchSide - \arrowHeight}    
    \foreach \x in {1,2,...,\gridWidth} {
            \ifnum \x > \thresholdImages
                \def\colorToken{blue!20}
            \else
                \def\colorToken{red!20}
            \fi
            \ifnum \x = \tokenDots
                \node at (\xDots, \yDots) {\dots};
            \else
                \draw[fill=\colorToken] (\x,0) rectangle ++(\rectWidth, \rectHeight);
            \fi
    }
    \draw[fill=green!20] (1, \yFalcon) rectangle ++(\falconWidth, \rectHeight);
    \node at (1 + 0.5*\falconWidth, \yFalcon + 0.5*\rectHeight) {\Large Falcon2-11B LLM (Chat Finetuned) $f_{LLM}(\cdot)$};

    \draw[fill=red!20] (1, \yProjector) rectangle ++(\projectorWidth, \rectHeight);
    \node at (1 + 0.5*\projectorWidth, \yProjector + 0.5*\rectHeight) {\Large Projector $f_{proj}(\cdot)$};

    \draw[->] (1 + 0.5*\projectorWidth,\yProjector+\rectHeight) -- (1 + 0.5*\projectorWidth,0) node[midway, right] {\Large $\bm{H}_p$};

    \draw[fill=red!20] (1, \yVisionEncoder) rectangle ++(\projectorWidth, \rectHeight);
    \node at (1 + 0.5*\projectorWidth, \yVisionEncoder + 0.5*\rectHeight) {\Large Vision Encoder $f_e(\cdot)$};    

    \draw[->] (1 + 0.5*\projectorWidth,\yVisionEncoder+\rectHeight) -- (1 + 0.5*\projectorWidth,\yProjector) node[midway, right] {\Large $\bm{H}_e$};

    \foreach \x in {1,2,...,\halfImages} {
        \ifnum \x = 1
            \node at (\x + 0.5*\patchSide, \yPatches) {\includegraphics[width=\patchSide cm, height=\patchSide cm,trim=0cm 11.25cm 11.25cm 0cm, clip]{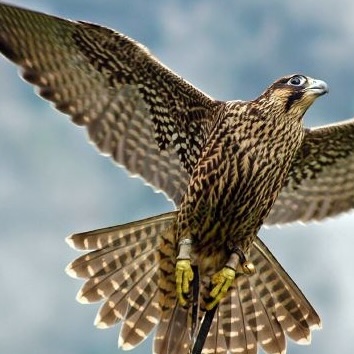}};
            \node at (\x + 0.5*\patchSide + \halfImages + 1, \yPatches) {\includegraphics[width=\patchSidecm, height=\patchSide cm,trim=8.75cm 0cm 2.5cm 11.25cm, clip]{figures/cropped_falcon.jpg}}; 
        \fi
        \ifnum \x = 2
            \node at (\x + 0.5*\patchSide, \yPatches) {\includegraphics[width=\patchSide cm, height=\patchSide cm,trim=1.25cm 11.25cm 10cm 0cm, clip]{figures/cropped_falcon.jpg}};
            \node at (\x + 0.5*\patchSide + \halfImages + 1, \yPatches) {\includegraphics[width=\patchSidecm, height=\patchSide cm,trim=10cm 0cm 1.25cm 11.25cm, clip]{figures/cropped_falcon.jpg}};        
        \fi
        \ifnum \x = 3
            \node at (\x + 0.5*\patchSide, \yPatches) {\includegraphics[width=\patchSide cm, height=\patchSide cm,trim=2.5cm 11.25cm 8.75cm 0cm, clip]{figures/cropped_falcon.jpg}};
            \node at (\x + 0.5*\patchSide + \halfImages + 1, \yPatches) {\includegraphics[width=\patchSidecm, height=\patchSide cm,trim=11.25cm 0cm 0cm 11.25cm, clip]{figures/cropped_falcon.jpg}};
        \fi
    }
    \node at (\xDots, \yPatches) {\dots};
    
    \node at (1 + 0.5*\projectorWidth, \yImage) {\includegraphics[width=\imageSide cm, height=\imageSide cm]{figures/cropped_falcon.jpg}};

    \draw[->] (1 + 0.5*\projectorWidth,\yImage+0.5*\imageSide) -- (1 + 0.5*\projectorWidth,\yPatches-0.5*\patchSide);

    \draw[fill=blue!20] (1 + \projectorWidth + \spacing, \yProjector) rectangle ++(\embWidth, \rectHeight);
    \node at (1 + \projectorWidth + \spacing + 0.5*\embWidth, \yProjector + 0.5*\rectHeight) {\Large Embedding};

    \draw[->] (1 + \projectorWidth + \spacing + 0.5*\embWidth,\yProjector+\rectHeight) -- (1 + \projectorWidth + \spacing + 0.5*\embWidth,0) node[midway, right] {\Large $\bm{H}_t$};

    \draw[fill=blue!20] (1 + \projectorWidth + \spacing, \yVisionEncoder) rectangle ++(\tokenizerWidth, \rectHeight);
    \node at (1 + \projectorWidth + \spacing + 0.5*\tokenizerWidth, \yVisionEncoder + 0.5*\rectHeight) {\Large Tokenizer};

    \node at (1 + \projectorWidth + \spacing + 0.5*\tokenizerWidth, \yPatches) {\large  \textit{User: What species is this bird? Falcon:}};

    \def\generatedTokensList{{The}, {bird}, {is}, {a}, {falcon}, {.},{\textbf{eos}}}
    \foreach \str [count=\x from \gridWidth] in \generatedTokensList {
        \ifnum \x = \gridWidth
            \draw[fill=gray!30] (\x,2*\spacing+2*\rectHeight) rectangle ++(\rectWidth, \rectHeight) node[pos=0.5] {\tiny \str} node[midway, left=4mm] {generated tokens};
        \else
            \ifnum \x > \totalTokens
            \else
                \draw[fill=gray!30] (\x,2*\spacing+2*\rectHeight) rectangle ++(\rectWidth, \rectHeight) node[pos=0.5] {\tiny \str};
            \fi
        \fi
    }
    \foreach \str [count=\x from \gridWidth] in \generatedTokensList {
    \ifnum \x < \totalTokens
        \pgfmathsetmacro{\actx}{\x+1}
        \draw[fill=gray!30] (\actx,\yTokens) rectangle ++(\rectWidth, \rectHeight) node[pos=0.5] {\tiny \str};
        \draw[fill=gray!30] (\actx,0) rectangle ++(\rectWidth, \rectHeight);
        \fi
    }

    \def\myList{{User:}, {What}, {spieces}, {is}, {this}, {bird} {?},{Falcon:}, {dummy}}
    \pgfmathsetmacro{\start}{\thresholdImages + 1}
    \foreach \str [count=\x from \start] in \myList {
        \ifnum \x > \gridWidth
        \else
            \draw[fill=blue!20] (\x,\yTokens) rectangle ++(\rectWidth, \rectHeight) node[pos=0.5] {\tiny \str};
        \fi
    }

    \draw[->] (1 + \projectorWidth + \spacing + 0.5*\tokenizerWidth,\yVisionEncoder+\rectHeight) -- (1 + \projectorWidth + \spacing + 0.5*\tokenizerWidth,\yTokens);
\end{tikzpicture}
}

\caption{Falcon2-11B VLM architecture. See \cref{falcon_vlm} for details.} 
\label{fig:vlm}

\end{center}
\end{figure}